\documentclass[10pt, conference, compsocconf, letterpaper]{IEEEtran}
\IEEEoverridecommandlockouts

\usepackage{cite}
\usepackage{amsmath,amssymb,amsfonts}
\usepackage{algorithm} 
\usepackage{algpseudocode} 
\usepackage{xcolor}
\usepackage{hyperref}
\usepackage{graphicx}
\usepackage{textcomp}
\usepackage{subfig}
\usepackage{multirow}
 \def\BibTeX{{\rm B\kern-.05em{\sc i\kern-.025em b}\kern-.08em
    T\kern-.1667em\lower.7ex\hbox{E}\kern-.125emX}}
\usepackage{enumitem}
\usepackage{colortbl}
\usepackage{soul}
\usepackage{graphicx} 

\newlist{inlineroman}{enumerate*}{1}
\setlist[inlineroman]{itemjoin*={{, and }},afterlabel=~,label=\roman*.}

\newcommand{\inlinerom}[1]{
\begin{inlineroman}
#1
\end{inlineroman}
}

\newlist{Inlineroman}{enumerate*}{1}
\setlist[Inlineroman]{itemjoin*={{, and }},afterlabel=~,label=\Roman*.}

\begin{document}

\setlist[enumerate]{nosep}
\setlist[itemize]{nosep}
\definecolor{mintgreen}{rgb}{0.6, 1.0, 0.6}
\definecolor{pastelviolet}{rgb}{0.8, 0.6, 0.79}
\definecolor{peridot}{rgb}{0.9, 0.89, 0.0}
\definecolor{richbrilliantlavender}{rgb}{0.95, 0.65, 1.0}
\definecolor{robineggblue}{rgb}{0.0, 0.8, 0.8}

\definecolor{green}{rgb}{0.1,0.1,0.1}
\newcommand{\done}{\cellcolor{teal}done}

\title{\textit{SynchroSim}: An Integrated Co-simulation Middleware for Heterogeneous Multi-robot System}

\author{\IEEEauthorblockN{Emon Dey$^1$, Jumman Hossain$^1$, Nirmalya Roy$^1$, Carl Busart$^2$}
\IEEEauthorblockA{	
$^1$Center for Real-time Distributed Sensing and Autonomy}
\IEEEauthorblockA{$^1$Department of Information Systems,
University of Maryland, Baltimore County, USA}
\IEEEauthorblockA{$^2$DEVCOM Army Research Lab, USA}
\IEEEauthorblockA{$^1$\{edey1, jumman.hossain, nroy\}@umbc.edu} $^2$carl.e.busart.civ@army.mil}

\maketitle
\begin{abstract}
With the advancement of modern robotics, autonomous agents are now capable of hosting sophisticated algorithms, which enables them to make intelligent decisions. But developing and testing such algorithms directly in real-world systems is tedious and may result in the wastage of valuable resources. Especially for heterogeneous multi-agent systems in battlefield environments where communication is critical in determining the system's behavior and usability. Due to the necessity of simulators of separate paradigms (co-simulation) to simulate such scenarios before deploying, synchronization between those simulators is vital. Existing works aimed at resolving this issue fall short of addressing diversity among deployed agents. In this work, we propose \textit{SynchroSim}, an integrated co-simulation middleware to simulate a heterogeneous multi-robot system. Here we propose a velocity difference-driven adjustable window size approach with a view to reducing packet loss probability. It takes into account the respective velocities of deployed agents to calculate a suitable window size before transmitting data between them. We consider our algorithm specific simulator agnostic but for the sake of implementation results, we have used Gazebo as a Physics simulator and NS-3 as a network simulator. Also, we design our algorithm considering the Perception-Action loop inside a closed communication channel, which is one of the essential factors in a contested scenario with the requirement of high fidelity in terms of data transmission. We validate our approach empirically at both the simulation and system level for both line-of-sight (LOS) and non-line-of-sight (NLOS) scenarios. Our approach achieves a noticeable improvement in terms of reducing packet loss probability ($\approx$11\%), and average packet delay ($\approx$10\%) compared to the fixed window size-based synchronization approach.
\end{abstract}
\section{Introduction}
In the field of contemporary robotics, multi-agent systems are set to play a vital part. Due to their ability of forming large interconnected networks with coordination among agents make them an integral part in a variety of robotic applications \cite{gomes2018co, ahn2018reliable}. For example, Unmanned Aerial Vehicle (UAV) systems are being increasingly used in a broad range of applications requiring extensive communications, either to collaborate with other UAVs \cite{sarkar2021uav} with each other or with Unmanned Ground Vehicles (UGV) \cite{acharya2020cornet, moon2020gazebo}. Specifically in battlefield scenarios where the presence of heterogeneous aerial and ground vehicles coordinating with each other is going to be an essential feature in the future. The mutual information transfer between Unmanned Aircraft Systems (UAS), and ground robots can help to make intelligent decisions in a critical time.

Synchronized communication between UAVs and UGVs can aid in developing situation awareness in the battlefield among each deployed device, and also help execute any specific command through them. Executing such systems directly in the actual environment may bring in harmful consequences as they necessitates the extensive fine-tuning of algorithm parameters \cite{calvo2021ros}. As a result, it is required to simulate the system beforehand using proper technologies in order to establish a baseline of confidence. Being motivated by this scope, research has been started on simulating such an environment to estimate the probable nature of robots before going to actual deployment. The main bottleneck in this endeavor lies in the necessity of synchronizing two different simulators having disparate operating principles \cite{baidya2018flynetsim, staranowicz2011survey}. One of these is physics simulators that account for replicating the interaction between physical robots and their operating environment. On the other hand, network simulators try to estimate the deployed agents' communication performance over the network.

So, the primary goal should be to connect the two domains by using existing open-source tools to record closed loop simulation. This is because multi-agent systems are strenuous to build; easy and coordinated communication across diverse technologies can make this process somewhat less difficult. Some of the existing works have already addressed those challenges. For example, FlyNetSim \cite{baidya2018flynetsim}, ROS-NetSim \cite{calvo2021ros}, CORNET \cite{acharya2020cornet}, CPS-Sim \cite{suzuki2018cps}, RoboNetSim \cite{kudelski2013robonetsim} are some of such kinds of works implemented on AirSim \cite{shah2018airsim}, ARGoS \cite{pinciroli2012argos}, Gazebo \cite{Psimulator} as physics simulator and OMNeT++ \cite{varga2010omnet++}, NS-2 \cite{NS-2}, NS-3 \cite{riley2010ns}, Mininet \cite{Mininet} as network simulator. As a whole, some of the major drawbacks existent is those works are: \inlinerom{\item compatible with either UAVs or UGVs, not heterogeneous systems, \item causes low co-simulation speed, \item gives rise to float-point arithmetic error, \item difficult to set up proper window size for diverse multi-agent setup \item loses synchronization when the simulators relative speed varies.}

With a view to developing a suitable ROS-compatible synchronizing middleware for the aforementioned scenario, we propose \textit{SynchroSim}. It takes into account the number of agents deployed within a certain cluster and can select the most suitable sliding window while sending data among agents. For simulation experiments, we use two distinct open-source simulation engines, Gazebo and NS-3. Gazebo uses the computer's system clock as the simulation time, while NS-3 presents the process as a distinct sequence of time events. Additionally, we observe the performance of our algorithm upon deploying on a cluster consisting of the real world UAVs and UGVs. To the best of our knowledge, it is the first endeavor pointing out the impact of synchronization middleware on a heterogeneous multi-agent system considering battlefield as application scenario. The major contributions we claim here are:
\begin{itemize}
    \item {\it Co-simulation of heterogeneous multi-agent systems i.e., UGV (ground robots), UAV (drones), etc. and preserving synchronization among them:} We propose a simulator independent co-simulation setting for multi-agent heterogeneous environment. Furthermore, we extend our simulation work into real world robots to validate the actual deployment performance.
    \item {\it Improvising sliding window-based synchronization scheme for diverse multi-agent system:} We offer an application specific modification of sliding window based synchronizing middleware. We name our approach as \textit{SynchroSim} which can vary the window size considering the velocity difference of the agents for the sake of synchronization among them with better communication performance.
    \item {\it Empirical evaluation considering different application scenarios:}
    We present our experiment, taking into account both line-of-sight (LOS) and non-line-of-sight (NLOS) communication scenarios on the basis of probability of packet loss, and average packet delay. We have employed Gazebo as Physics simulator and NS-3 as network simulator to report our experimental results. Experimental results show that our approach works better in comparison with the traditional fixed window based method in ensuring fewer packet losses (on average 10\% improvement) even in challenging NLOS environment with heterogeneous agents.
\end{itemize}

\section{Related work}
In this section, we will briefly outline the work that has been done in relation to the various aspects of our system.
\subsection{Simulation Tools}
While a comprehensive assessment of currently utilized simulators is beyond the scope of this paper, we highlight a few recent noteworthy physics and network simulators that are most relevant to our setting and have had a significant influence on this study. 

Aiming to bridge the gap between simulation and reality, AirSim \cite{shah2018airsim} is an open-source platform that is being developed to assist in the development of autonomous vehicles. AirSim provides high-fidelity physical and visual simulation that enables the rapid generation of massive amounts of training data for the development of machine learning models. It's API design enables algorithms to be developed against a simulator and then deployed unchanged on real vehicles. ANVEL \cite{durst2012real, fields2016simulation} provides such a toolkit by integrating popular graphical representation approaches, such as those used in video games, with physically based sensor and UGV platform models. While both AirSim and ANVEL have major simulation capabilities, difficulty arises when creating large-scale complex visually rich environments that are more realistic in their representation of the real world, and they have fallen behind various advancements in rendering techniques made by platforms such as Unreal engine or Unity \cite{Usimulator}. We prefer to employ Gazebo in our communication-realistic scenario because of its superior realism. Gazebo \cite{moon2020gazebo} includes a modular design that enables the usage of various physics engines, sensor models, and the creation of 3D worlds to be implemented. 

In the case of network simulators, OMNeT++ \cite{varga2010omnet++} is a discrete event network simulation framework that is object-oriented and modular in design. Additionally, parallel distributed simulation is supported by OMNeT++ and inter-participant communications can be accomplished through a variety of methods. A network emulation program, mininet \cite{Mininet}, allows users to create a realistic virtual network on a single computer by running genuine kernel, switch and application code on the network emulator. However, we choose a state-of-the-art event based simulator NS-3 \cite{Mininet}, in which the scheduler typically performs the events in a sequential manner without syncing with an external clock. The NS–3 simulator includes representations of all of the network models that make up a computer network including network nodes, network devices, communication channels, communication protocols, protocol headers, and network packets. 

\subsection{Synchronization Methods}

At this point, we provide a summary of existing synchronization methods.

CORNET \cite{acharya2020cornet} is a variable-stepped multi-robot system simulation framework that blends physics and network simulators. CORNET ensures that only one event process can be running at a time, and a global event scheduler maintains a list of all the events from both simulators and schedules. CORNET has the most significant downside is that it has the potential to cause float-point arithmetic error. CORNET 2.0 \cite{acharyacornet2} extended the implementation of CORNET to make it applicable to any robotic framework and the scalability to manage an increasing number of robots has been demonstrated. FlyNetSim \cite{baidya2018flynetsim} maintains a time-stepped-schedule mechanism, however; simulated network events must be buffered until the next sample time if a network simulator runs faster than the physical simulation. Using real-world UAVs and sensors in a simulated complicated environment is possible with the support of emulation mode in FlyNetSim where the network simulator allows a real UAV to communicate with external or simulated resources. 

On the other hand, CPS-Sim \cite{suzuki2018cps} operates on the basis of a global event-driven system where the server is forced to reproduce the exact same time-steps. Global scheduler-processed events resolve the issue inherent in the time-stepped method, albeit at the expense of overall co-simulation speed. ROS-NetSim \cite{calvo2021ros} is a sliding window based technique that keep track of and record network events during the course of the window period, and then enable the network simulator to step up to and through the end of the window, but it is difficult to configure the appropriate window size for multi-agent systems. In this consideration, we devised the sliding window method for our heterogeneous multi-agent system. Additionally, we have added a comparison table.
\begin{table*}[htbp]
\centering
\caption{\label{tab:related}Overview of existing synchronization methods.}
\begin{tabular}{|l|l|l|l|l|}
\hline
\multicolumn{1}{|c|}{\textbf{Middleware}} & \multicolumn{1}{c|}{\textbf{Synchronization Method}} & \multicolumn{1}{c|}{\textbf{Principle}}                                                                                                                                                                                                   & \multicolumn{1}{c|}{\textbf{Drawback}}                                                                                                                                                                                  & \multicolumn{1}{c|}{\textbf{Compatibility}} \\ \hline
Ros-NetSim \cite{calvo2021ros}                                & Sliding Window                                       & \begin{tabular}[c]{@{}l@{}}Capture and track network events\\ over the window period and allow\\ the network simulator to step up to\\  the end of the window.\end{tabular}                                                               & \begin{tabular}[c]{@{}l@{}}Difficult to set up proper window \\ size for multi-agent systems.\end{tabular}                                                                                                              & ROS1, UAV \& UGV                            \\ \hline
CORNET \cite{acharya2020cornet}                                   & Variable-stepped                                     & \begin{tabular}[c]{@{}l@{}}A global event scheduler maintains\\ the list of all the events from both \\ the simulators and schedules \\ according to their timestamps \\ allowing only one event process\\ to run at a time.\end{tabular} & Float-point arithmetic error                                                                                                                                                                                            & ROS1 \& UAV                                 \\ \hline
FlyNetSim \cite{baidya2018flynetsim}                                 & Time-stepped with scheduler                          & \begin{tabular}[c]{@{}l@{}}Common sampling period to be\\ used by both simulators.\end{tabular}                                                                                                                                           & \begin{tabular}[c]{@{}l@{}}If the network simulator runs \\ faster than the physics simulator,\\ the network events must be buffered\\ in a cache and wait to be processed\\ until the next sampling time.\end{tabular} & ROS1 \& UAV                                 \\ \hline
CPS-Sim \cite{suzuki2018cps}                                  & Global event driven                                  & \begin{tabular}[c]{@{}l@{}}The server is forced to reproduce\\ the exact same time-steps.\end{tabular}                                                                                                                                    & \begin{tabular}[c]{@{}l@{}}Global scheduler processed events\\ overcomes the problem that occurs\\ in time-stepped method but limits\\ the overall co-simulation speed.\end{tabular}                                    & Not ROS based                               \\ \hline
\end{tabular}
\end{table*}
\noindent(Tab. I) for different synchronization mechanisms to get a better idea about their capabilities and limitations.

In summary, UAV and UGV components are being simulated with 3D visualization using Gazebo, while the network infrastructure is being provided by NS-3 and middleware is being developed for the creation of an inter-simulation data-path with time and position synchronization at both ends using our co-Simulation of robotic networks.
\section{Overview}
\begin{figure}[htbp]
\centering
\includegraphics[scale=0.8, width=\linewidth]{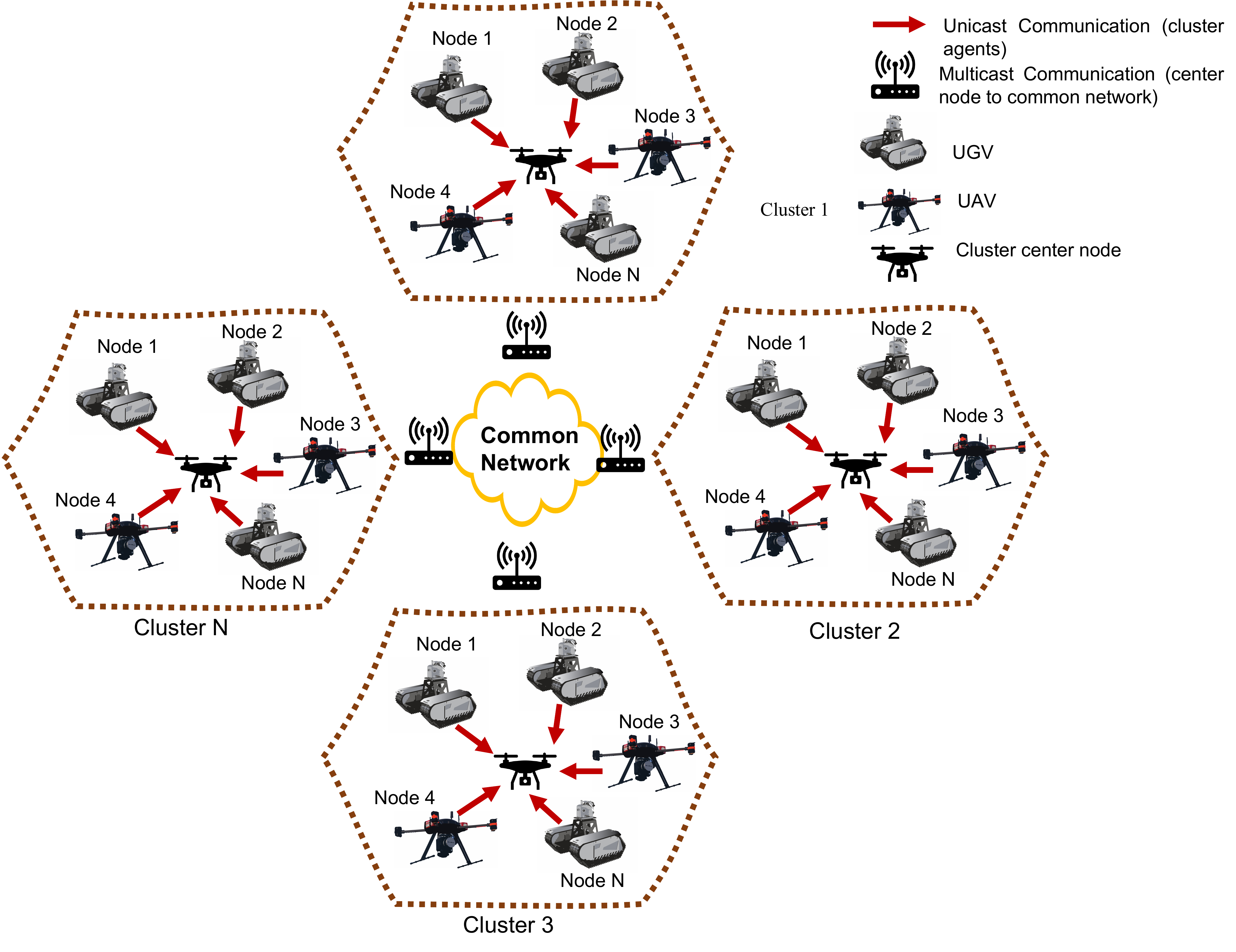}
\caption{A sample multi-master communication framework.}
\label{clus}
\vspace{-2ex}
\end{figure}

\begin{figure}[htbp]
\centering
\includegraphics[scale=0.8, width=\linewidth]{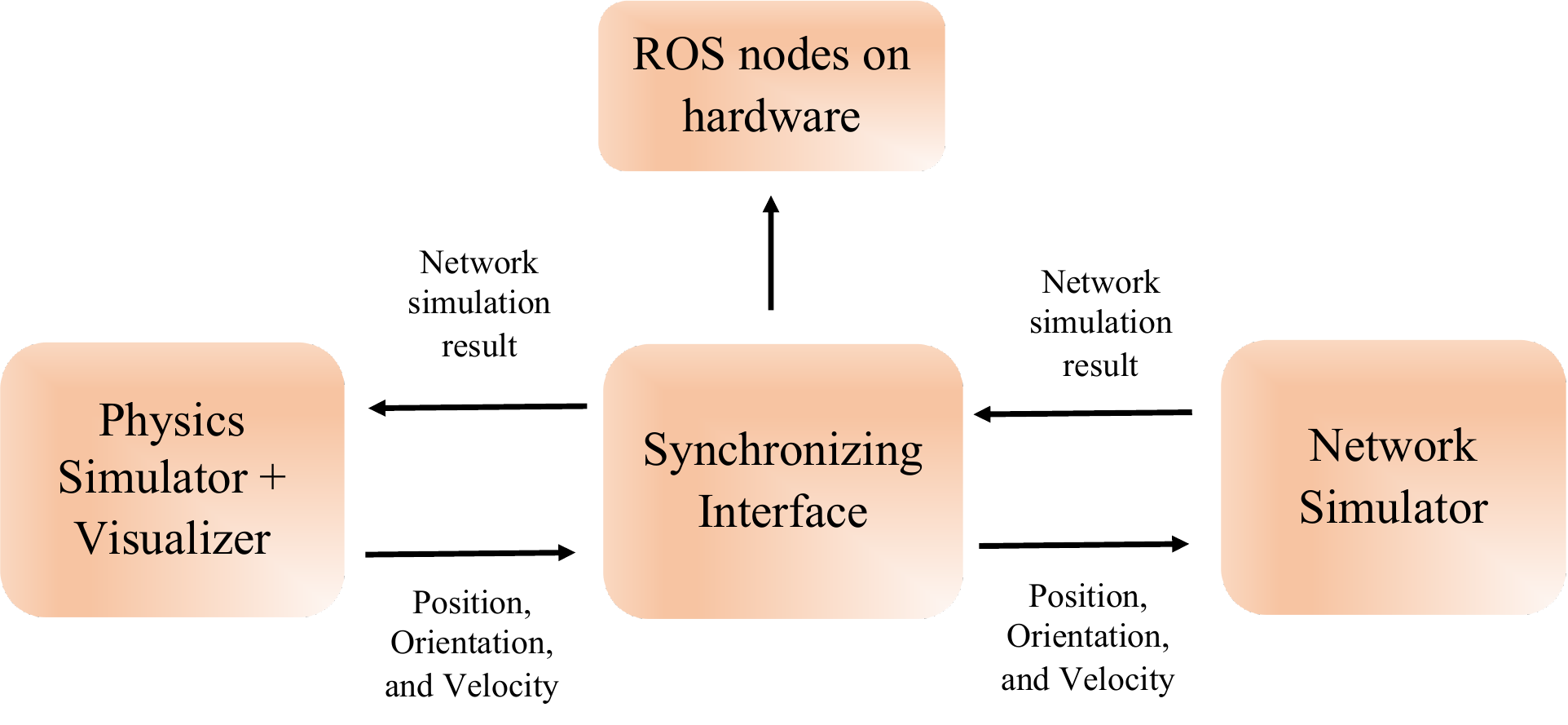}
\caption{A probable solution to the synchronization issue between heterogeneous simulators.}
\label{tech}
\vspace{-2ex}
\end{figure}

In modern battlefield scenarios, where intelligent and different sensor-equipped robots are envisioned to co-exist with soldiers. Those robots can collect data with the integrated sensors and can take essential decisions on their further step through analyzing the collected data. This can aid in increasing situational awareness if the derived information can be exchanged with other deployed agents and with the base stations also. Such a scenario is illustrated in fig \ref{clus}. 
Here, the total deployed agents are divided into several clusters. Each cluster has a cluster head. The cluster heads can communicate with the agents within the cluster and also among themselves. Furthermore, all of them can report to a central base station. Now, before going to the deployment with actual robots of such systems, it is imperative to simulate such scenarios to get an idea about the probable behavior after deployment.
For this reason, a combination of a Physics simulator and a Network simulator is needed to emulate such a communication scenario in an acceptable manner. But synchronization problem arises when these two kinds of simulators are asked to collaborate. As a solution to this, a synchronization middleware can be used to bridge that gap. A working flowchart of such systems is shown in fig \ref{tech}. 
This system takes input from the Physics simulator, the middleware then aligns those data with a networking event. The Network simulator then measures the communication performance and those results can be displayed on the physics simulator interface if needed. Moreover,  at the time of actual deployment, it is possible to run the ROS nodes of synchronization middleware on the integrated computing devices of the robots.
\section{Methodology} \label{method}
In this section, we will walk you through the details of our proposed algorithm and its working procedure on the selected application.
\subsection{\textit{SynchroSim} working Procedure}
As described in the earlier sections, our work focuses mainly on the synchronization aspect of heterogeneous multi-agent setup. Our proposed algorithm is based upon the sliding window-based synchronization approach proposed in \cite{calvo2021ros}. The pivotal design parameter while working with a multi-agent system is the choice of window size. Fixed window size at all points will not applicable in such a scenario as different agents may work better with different radio frequencies and also there can be disparities in terms of hardware specifications. \textit{SynchroSim} takes into account the total number of agents present for a  specific scenario and starts with a manually initiated value for window size. In the event of the data transfer, it selects the respective publisher and subscriber for that event. After that, it rigorously checks the velocity difference of the participating agents to calculate a suitable window size for a specific communication event. This approach becomes more evident when it comes to the transmission scenario between a ground and an aerial vehicle. To maintain the transmitted information level at a satisfactory point, we chose to utilize packet loss probability value, one of the most important parameters to monitor while sending valuable information, in the operating moment of \textit {SynchroSim}. The following equation is employed to inherently calculate packet loss probability and report within a synchronization window:\\
Packet loss probability,
\begin{equation}
{ }_{k}^{i} L p=1-\frac{{ }^{i} N_{s b}}{{ }_{k}^{i} N_{p b}}
\end{equation}
Here, \({ }_{k}^{i} N_{s b}\) is the number of data packets delivered to the subscriber and \({ }_{k}^{i} N_{p b}\) is the number of data packets transmitted from the publisher.
More technical details on the process can be found in Algorithm \ref{algo:1}.
\begin{algorithm}
	\caption{Multi-agent synchronization algorithm with adjustable window}
	\begin{algorithmic}[1]
	\Require Total number of agents $D$, window size $w$, number of data packets $N$, velocity of agents $V$
	\Ensure Synchronization between simulators and calculate packet loss probability $Lp$\\
	\textbf{Initialize:} Publisher P and subscriber S agents where $P,S\in D$, begin time = 0, window size w \\
	\textbf{Start simulation:} Transmit data packet of selected topic from publisher\\
	    \textit{Update} with begin time t = 0
	    \If {running event found}
	    \State\textbf{Window adjustment:} Get the velocity of Publisher $V_p$ and Subscriber $V_s$ in meters/sec\\
    	Calculate the adjusted window, \begin{equation}
w_{a}=w+\left(V_{p}-V_{s}\right) / 1000
\end{equation}
	    \State\textbf {Synchronism Period:} Wait until begin time = t
	    \State\textbf{Report Lp:} Calculate packet loss probability for the synchronized event
	    \State\textbf{Report finished window:} Send \textit{Update} with end time = t
	    \State\textbf{Timestamp update:} \textit{Update} $t = t + w_a$ and request for next window
	    \EndIf
	\end{algorithmic} 
    \label{algo:1}
\end{algorithm} 
\subsection{Updating Physics and Network Simulator}
At the beginning of the simulation, the Physics simulator determines two key pieces of information about the agents used for a specific communication round: the distance between the agents through positional coordinates and velocity. The network simulator stores information on which communication scheme to use (we have chosen a TCP/IP-based approach over UDP for reliability issues), the number of packets, packet length, and the IPs of both publisher and subscriber agents. Then the Physics simulator and network simulator are advanced with one initially fixed window size \textit{w}. Upon the advancement, the Physics simulator passes the distance and velocity information to the network simulator, and the network simulator calculates the packet loss probability. If the loss value is within a certain threshold, the result is sent to display. Otherwise, this information is reported back to the \textit{Synchrosim} module and the initial window size is adjusted according to the algorithm \ref{algo:1}. This process continues until a satisfactory result is achieved for this specific round, and then the initialization process starts over with a new set of agents.
\subsection{Publisher-Subscriber Architecture}
\begin{figure}[htbp]
\centering
\includegraphics[width=\linewidth]{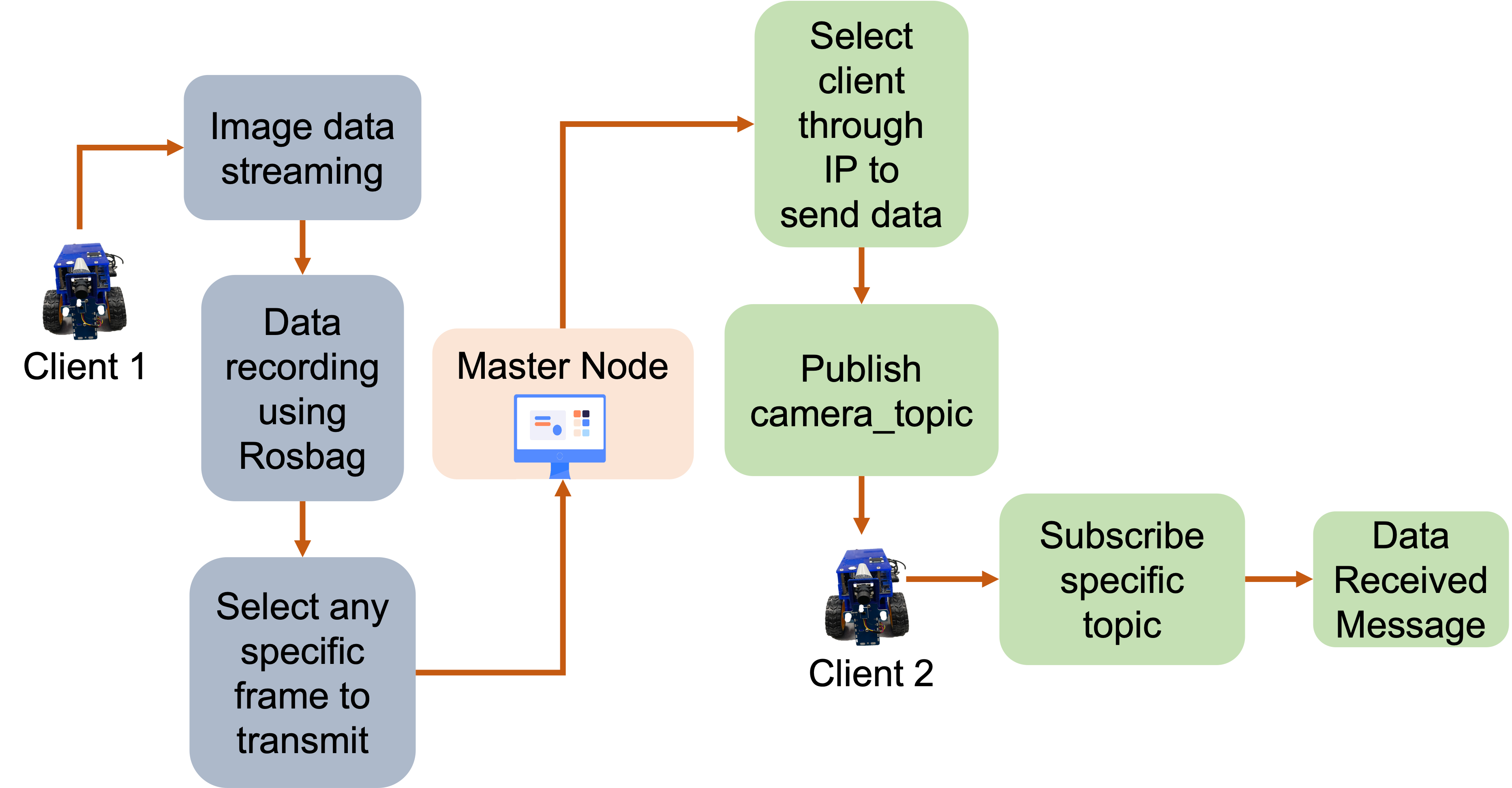}
\caption{Data transmission flowchart through the master node.}
\label{master}
\vspace{-2ex}
\end{figure}

\textit{SynchroSim} utilizes a publisher subscriber-based architecture to accomplice the data transmission task. We have considered image as our data type. A cluster was set up with a combination of a drone and two ground bots. Heterogeneity was maintained while choosing the ground bots and master-based ROS \cite{quigley2009ros} communication was implemented. In the case of communication through the master, all the agents were set to run within a certain area. The flowchart for this master-based setup is illustrated in fig \ref{master}. The drone was chosen as the master node and it contained the IPs of all the UGVs. Image data were streaming from all the UGVs. We have utilized the Rosbag concept to record the published messages and select any specific frame to transmit. Rosbag stores the information it is programmed to save with every timestamp. The images information can be extracted from that inventory. When an event was initiated, the selected image frame was sent to the master node. The master node then published the relevant ROS topic and any agent subscribed to that specific topic can be eligible to receive that image data. After the completion of this transmission process, the packet loss probability for that event was monitored.

\subsection{Setting up clusters and communication schemes}
\begin{figure}[htbp]
\centering
\includegraphics[width=\linewidth]{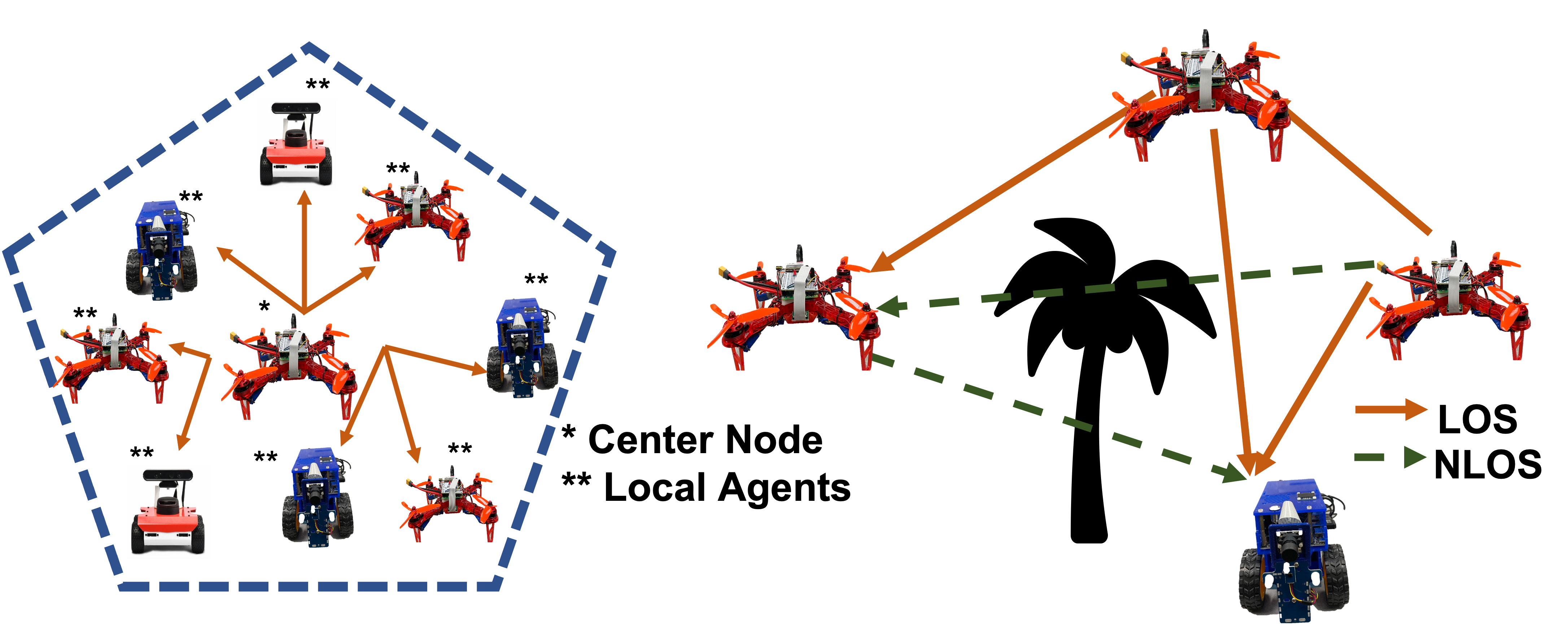}
\caption{A cluster consisting multi-agent heterogeneous systems (left), and Visual demonstration of LOS and NLOS communication (right).}
\label{cluster}
\vspace{-2ex}
\end{figure}

In the wireless communication domain, a cluster of robots is a popular setting to measure the inter-agents data transmission performance. In application scenarios involving contested environments such as modern battlefields where the flow of information among heterogeneous multi-robot systems is crucial to increase situational awareness. Among different types of cluster setup, the master-based system is being investigated intensively in recent ROS-based research. A sample master-based multi-agent cluster is illustrated in fig \ref{cluster}. In this work, we will also utilize this scheme.\\
To bolster the claim of a working synchronizing middleware it is imperative the set up detailed experimentation with two of the most used communication scheme: line-of-sight (LOS), non-line-of-sight (NLOS). To define those two schemes, when there is no obstruction in between the receiver and sender agents, the data transmission rate becomes higher and this scenario is known as LOS communication. It is the most desirable situation for wireless signal transmission. The NLOS scheme is the opposite of LOS communication and it is the most common one in the real-world environment. The signal transmission can be hindered by both natural and man-made objects. So, a better design of the data transmission scheme is necessary to avoid the loss of valuable information. A simple visual representation of LOS an NLOS system can be found in fig \ref{cluster}. For our experimentation, we have simulated both UGV and UAV with artificial environment settings containing both types of communication scenarios.

\section{Simulation Setup} \label{simulation}
\begin{figure*}[htbp]
  \centering
  \subfloat []{\includegraphics[height=5cm, width=0.48\textwidth]{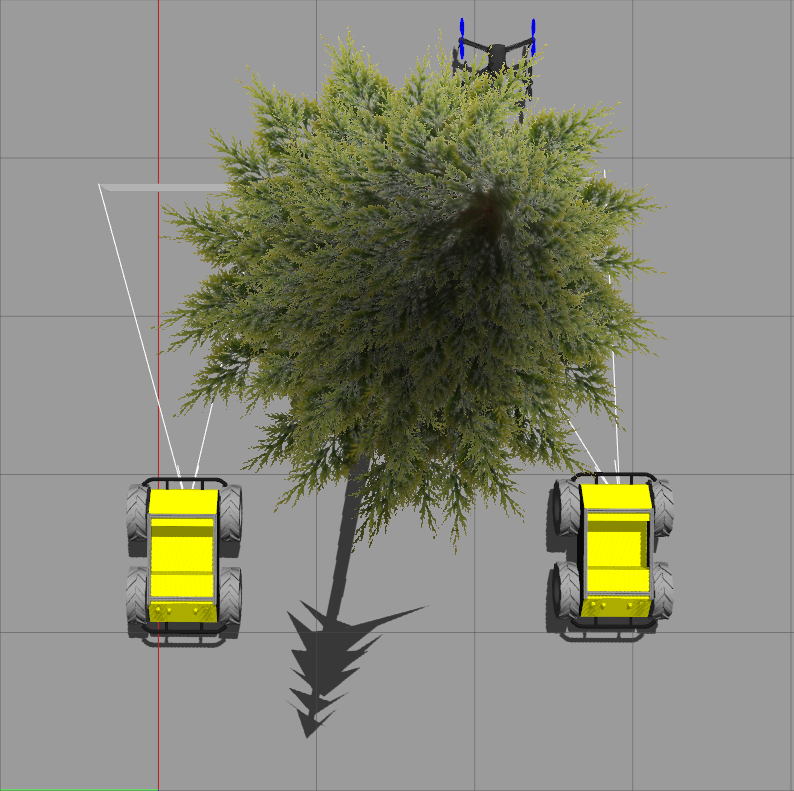}}
  \hfill
  \subfloat []{\includegraphics[height=5cm, width=0.48\textwidth]{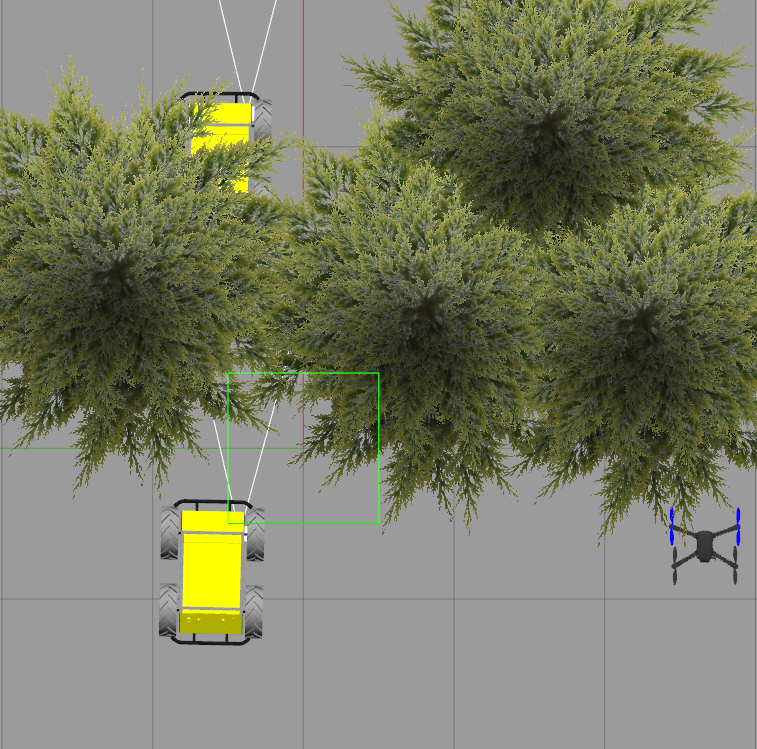}}
  \caption{Simulation environment design in Gazebo for (a) LOS, and (b) NLOS communication. The whole environment is 100m$\times$100m, and each grid stands for 20m$\times$20m.}
  \label{gazebo}
\vspace{-2ex}
\end{figure*}

To validate the working capability of \textit{SynchroSim}, we have arranged simulation scenarios for both LOS/NLOS channels. The environments are launched on Gazebo; the Physics simulator in our case. The dimension of the chosen environment is 100$\times$100 meters as shown in fig \ref{gazebo}. Where each of the grids signifies 20$\times$20 meters. The environment is designed to host both LOS/NLOS scenarios and for NLOS abstraction, the environment is populated with trees mostly. The choice of robotic agents is done in a heterogeneous fashion; contains both UGV and UAV. Iris drone with integrated camera as UAV, and two Husky robots are selected as UGV. One of the Husky robots is considered the master node. We have used the \textit{MAVROS} package to establish the connection between the Gazebo and the Iris drone, which runs on the \textit{MAVLink} autopilot setup and an integrated PX4 flight stack to maneuver the drone. For the Husky simulation, we modified \textit{Clearpathrobotics'} official GitHub implementation process according to our use case. For wireless communication medium, we have used IEEE 802.11 (Wi-Fi) interface. The agents are set to stream datapoints through TCP/IP link and the master node has the IP address of each of the deployed agents. The integrated camera of those agents is used as the primary sensor and the streaming image frames are resized into 32$\times$32 before treating them as camera topic of the ROS system. To measure the communication performance, we have chosen one of the state-of-the-art network simulators, NS-3 which is compatible with the selected wireless stack. To integrate both Gazebo and NS-3, \textit{SynchroSim} is deployed as synchronizing middleware. The choice of window size in \textit{SynchroSim} is dependent upon the respective velocity of the agents. To execute this scenario, the velocity of the agents are varied accordingly. The initial window size is set to 1mS and is adjusted with Algorithm \ref{algo:1}. 
\begin{figure*}[htbp]
  \centering
  \subfloat []{\includegraphics[width=0.48\textwidth]{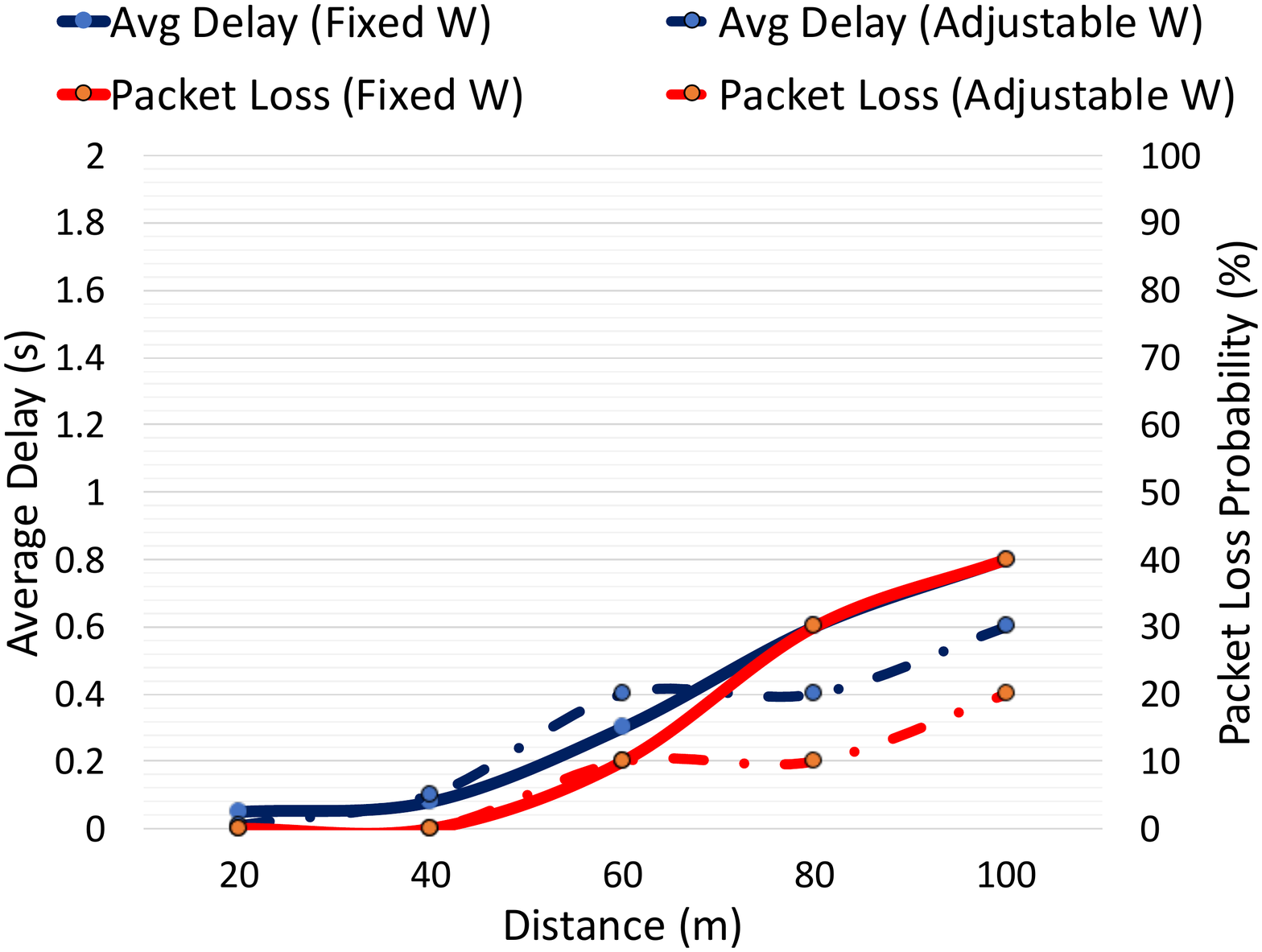}}
  \hfill
  \subfloat []{\includegraphics[width=0.48\textwidth]{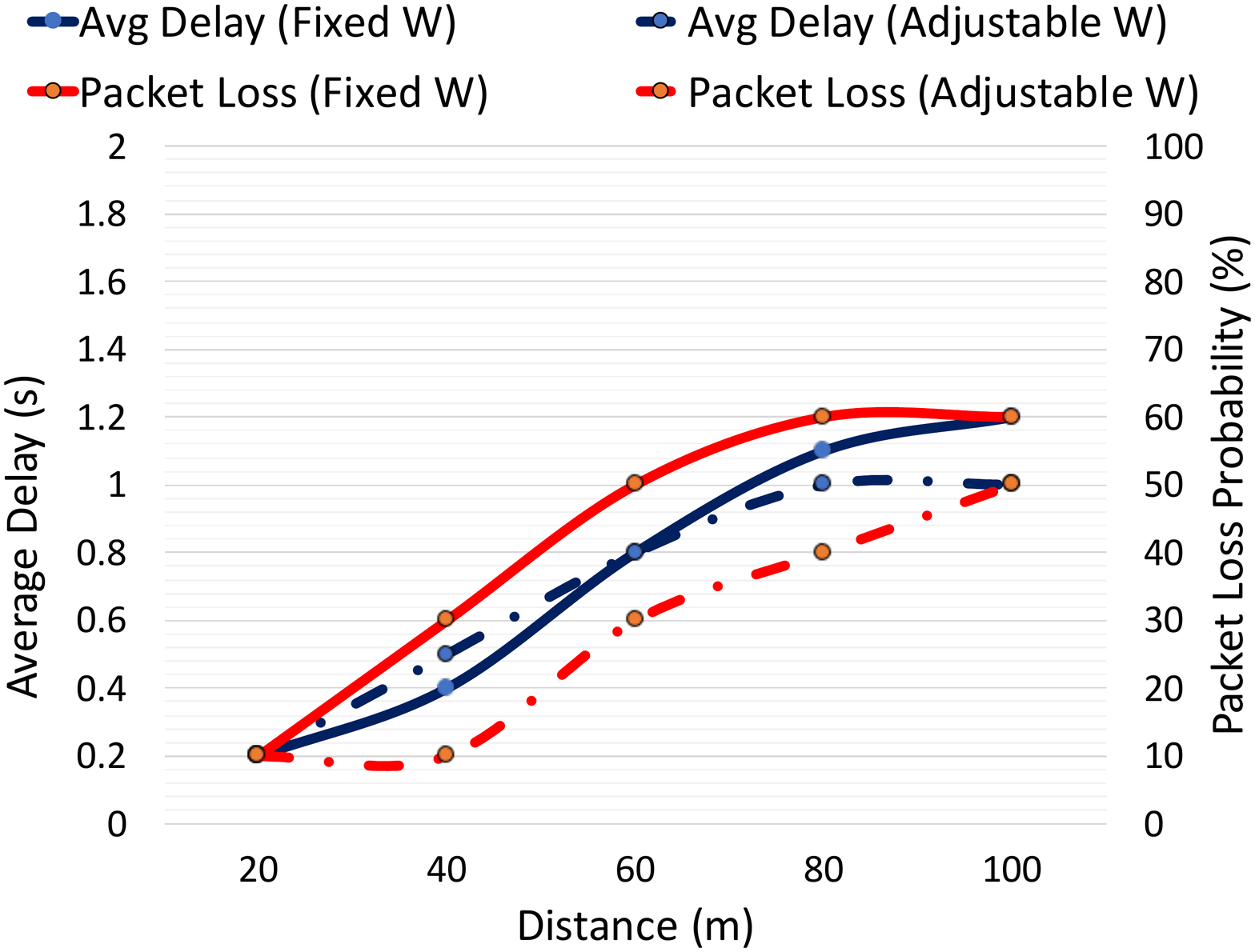}}
  \caption{Result comparison between fixed window and adjustable window based approaches on both (a) LOS, and (b) NLOS communication scheme using average delay (s) and packet loss probability (\%) matrices considering a UGV to UGV communication scenario. The fixed window size chosen here is 1ms and the average delay is reported on a batch of 10 packets. Blue lines stand for the average delay values and orange lines are for percentage of packet loss. In both cases the dotted lines signify the results of our proposed method.}
  \label{ugug}
\vspace{-2ex}
\end{figure*}
\begin{figure*}[htbp]
  \centering
  \subfloat []{\includegraphics[width=0.48\textwidth]{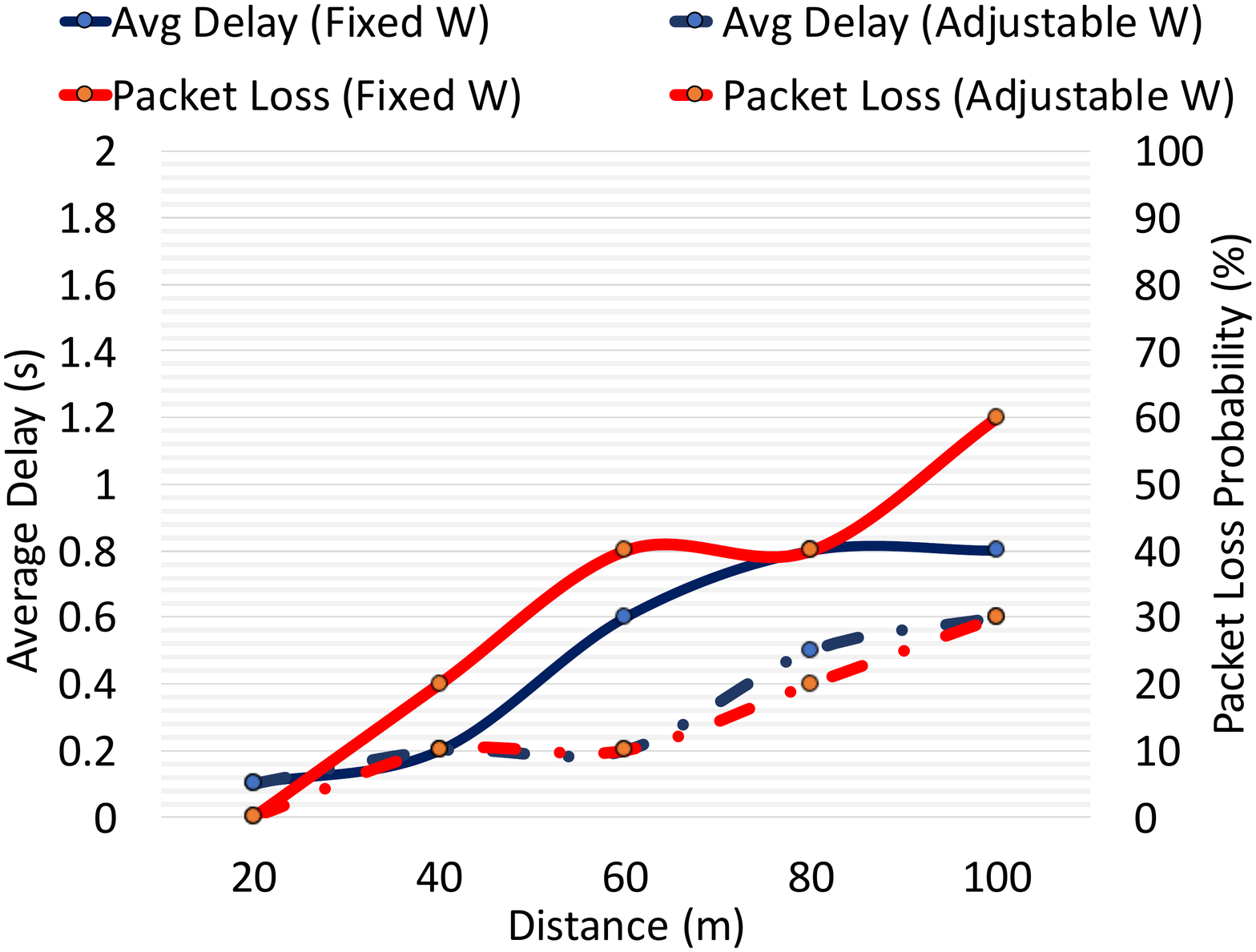}}
  \hfill
  \subfloat []{\includegraphics[width=0.48\textwidth]{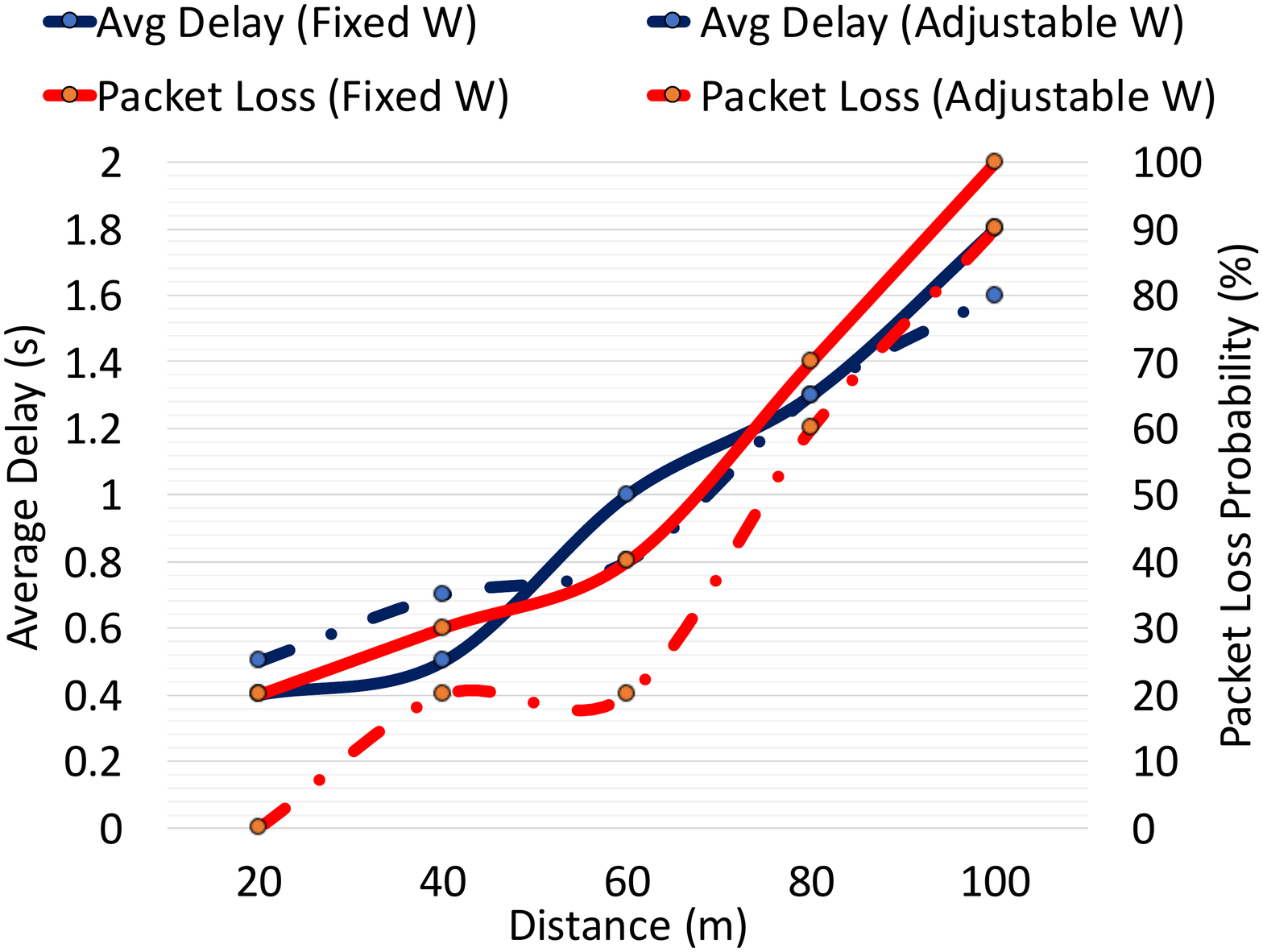}}
  \caption{Result comparison between fixed window and adjustable window based approaches on both (a) LOS, and (b) NLOS communication scheme using average delay (s) and packet loss probability (\%) matrices considering a UGV to UAV communication scenario. Same measurement paradigm and legend conventions as fig \ref{ugug} are used for this figure also.}
  \label{ugua}
\vspace{-2ex}
\end{figure*}
For the LOS scenario, the agents are deployed in an environment without obstruction as shown in fig \ref{gazebo}. The UGVs and the UAV are simulated with varying speeds and the communication performance is measured with the LOS abstraction is satisfied. The performance matrices are the average delay and percentage of packet loss as stated in section \ref{method}. The results are reported with the change in distance between the agents. Also, two different communication scenarios are considered which we define as UGV to UGV and UAV to UGV communication. The distance values reported while reporting the experimental results are derived from the positional coordinates of the agents. The whole simulation area is segmented into symmetrical blocks and the synchronization operation is operated when the agents move to some certain points of the environment.

In the case of NLOS abstraction, the agents are set to run within an object-populated environment. The signal strength is hypothesized to be hampered under this circumstance. The same reporting paradigm and communication scenarios are used for this case also. 

During UGV to UGV communication where the relative velocity difference is lower, the agents are noticed to sustain the data quantity during transmission for about 40m as shown in fig \ref{ugug}. The average delay (average value of delays during 10 image frame transmission) is also below 0.1s in this case. The velocity information of two selected agents for a communication incident is extracted through subscribing to $/gazebo\_states/twist$ topic. The baseline fixed window-based approach and the proposed adjustable window method are working at par for LOS communication up to this point. But fixed window-based approach begins to lose valuable information drastically after this point subsequently giving rise to delay. A slight adjustment in the sliding window value (0.1ms in this case) is seen to contribute to almost 20\% improvement in retrieving information for LOS and at least 10\% for NLOS condition when the agents are the furthest distance apart in this simulation (100m). Also, the transmission is becoming faster by a similar percentage compared to the baseline. 

While experimenting with UGV to UAV, severe performance degradation is seen for NLOS communication, illustrated in fig \ref{ugua}. To be specific, almost all the data are lost when the distance is 100m between the agents for the fixed width method. This scenario also affects the proposed adjustable window approach (0.3ms increase) which leaves a point of improvement. Apart from this specific point, the adjusted window showcases significant improvement for both LOS and NLOS communication. Noticeably, the packet loss probability is reduced by almost 30\% when we consider a LOS scenario and the agents are furthest apart.

\section{System Implementation}
In this section, we describe the devices used for our hardware-level implementation, procedure details, and performance evaluation.\\
With a view to validating the application compatibility of our proposed algorithm on actual systems, we have conducted a system-level implementation and evaluation. We have chosen ROS-compatible Duckiebots to act as our deployed agents. A brief description of the configuration of such robots are given below:

\begin{figure*}[htbp]
  \centering
  \subfloat []{\includegraphics[width=0.23\textwidth]{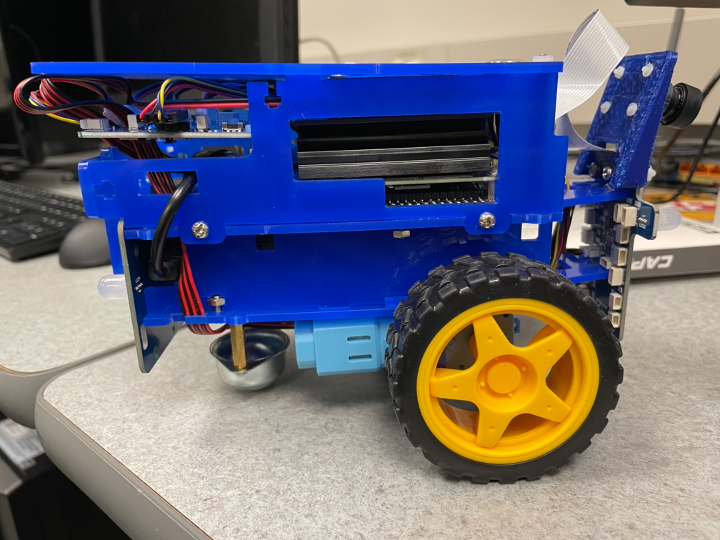}}
  \hfill
  \subfloat []{\includegraphics[width=0.23\textwidth]{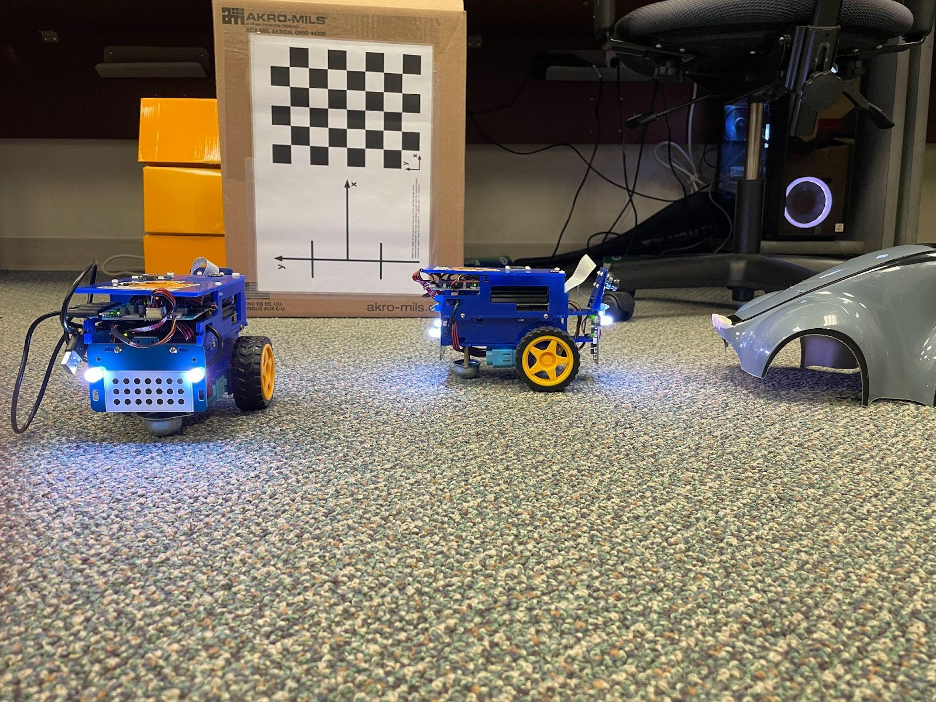}}
  \hfill
  \subfloat []{\includegraphics[width=0.23\textwidth]{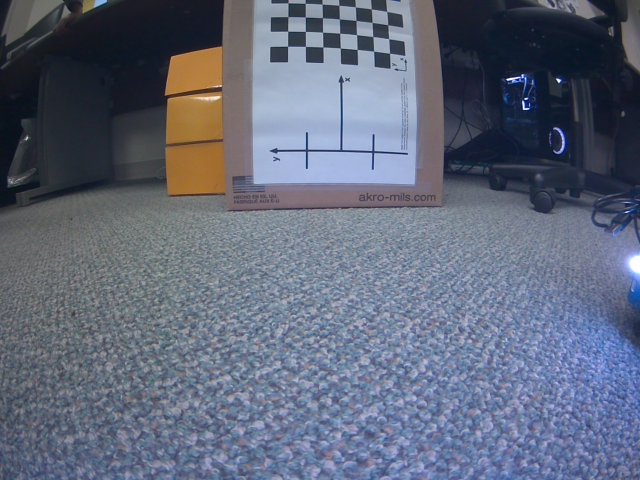}}
  \hfill
  \subfloat []{\includegraphics[width=0.23\textwidth]{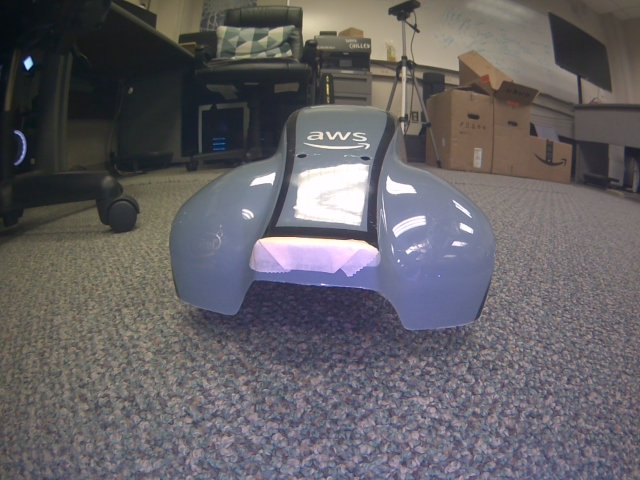}}
  \caption{(a) A fully assembled Duckiebot (UGV), (b) Image data transmission between two Duckiebots using a master based system in a laboratory environment. Sample sensor output of (c) agent 1, and (d) agent 2 within the experimental setup.}
  \label{labro}
\vspace{-2ex}
\end{figure*}
\textbf{Duckiebot:}\\
The Duckiebot is an autonomous platform developed for research purposes and convenient for studying complex real-world problems. The bot used in our work has sensors like a front-facing camera, programmable LED, IMU; camera inputs are used for our implementation. The camera used in this specific Duckiebot version is a 5MP, 1080p camera with a wide view range of 160 degrees. We have assembled the robot from scratch to make sure it is equipped with all the necessary components we need to operate the experiment. A fully assembled Duckiebot is illustrated in fig \ref{labro}. Its onboard processor of it is Nvidia Jetson Nano (2GB). Docker containers are built to run the ROS nodes and all the necessary scripts are written on Python.


As stated in the earlier sections, we have implemented a master-based multi-agent communication scheme. This implementation operation is carried out as a corroboration attempt of the simulation results obtained and described in section \ref{simulation}. An Ubuntu desktop was set up as the master node and two Duckiebots were clients. The master node will have the TCP of each agent and can receive or send information to any cluster agents based on the requirement. This implementation is carried out through a Publisher-Subscriber approach based on ROS Melodic. The experimentation is done in a laboratory environment. The experimental setup along with the sample camera outputs (primary sensor) of the two Duckiebots are illustrated in fig \ref{labro}. The Duckiebots are set to roam around in an approximately 2m x 2m area and some sample objects are placed within it. The robots are continuously capturing image frames and as soon as an object is detected by one bot, it will send the corresponding frame to the master node. The master node will then send this information to the other bot. A visualizer is designed to check the completion of data transmission and also the communication performance packet delay and packet loss probability. The synchronization algorithm was deployed on the docker container to run on behind when it is needed to display the forwarded data from the clients and also the performance metric values. The data used for transmission is a gray scale image which is resized in 32$\times$32 before sending. Under the aforementioned laboratory setup, \textit{SynchroSim} works seamlessly in a confined small scale setting, ensuring negligible delay and no packet loss for both LOS and NLOS abstraction. This successful demonstration makes us hopeful to achieve satisfactory performance with our ongoing deployment endeavor in wild contested environment.

\section{Conclusion and future work}
In this work, we have presented \textit{SynchroSim}, a middleware for co-simulation of heterogeneous multi-robot systems. \textit{SynchroSim} can adjust the size of the window based on the speed differences between the agents in order to provide better synchronization and communication. Even in the challenging non-line-of-sight (NLOS) environments with heterogeneous agents, experimental data show that our solution outperforms the standard fixed window based strategy in terms of ensuring fewer packet losses (on average 10\% improvement). Furthermore, we have also arranged a small scale cluster in a laboratory environment with real world UGVs to get an idea of the applicability of our proposed synchronizing approach when it comes to real terrain. In this work, we have demonstrated our synchronizing method on master-based communication system. However, one interesting extension can be implementing a masterless communication to ensure more data security; one of the fundamental requirements of battlefield scenarios. We plan to explore the facility offered by ROS2 on top of our current system with the help of ROS bridge to make the robots enable to use the Data Distribution Service (DDS).
\section*{Acknowledgment}
This work has been supported by U.S. Army Grant \#W911NF2120076.
The authors would also like to thank Dr. Kasthuri Jayarajah and Dr. Aryya Gangopadhyay for their constructive feedback on this work,  Jonathan Harwood for setting up the simulation and Sreenivasan Ramasamy Ramamurthy for helping with Duckiebot assembly.

\bibliographystyle{unsrt}
\bibliography{main}
\end{document}